\documentclass[a4paper, 10pt, conference]{ieeeconf}      

\IEEEoverridecommandlockouts                              
\usepackage{graphicx} 
\usepackage{subfigure}
\usepackage{xcolor}
\usepackage{multirow}
\usepackage{siunitx}
\usepackage{booktabs}
\usepackage{tabularx}
\usepackage{fullwidth}

\title{\LARGE \bf
An Attentive Sequence Model for Adverse Drug Event Extraction\\ from Biomedical Text
}

\author{Suriyadeepan Ramamoorthy \hspace{1em} Selvakumar Murugan \\
SAAMA AI Research Lab \\
Chennai, India\\
{\tt\small \{suriyadeepan.ramamoorthy, selvakumar.murugan\}@saama.com}
}

\begin{document}

\maketitle
\thispagestyle{empty}
\pagestyle{empty}

\begin{abstract}

Adverse reaction caused by drugs is a potentially dangerous problem which may lead to mortality and morbidity in patients. Adverse Drug Event (ADE)  extraction is a significant problem in biomedical research. We model ADE extraction as a Question-Answering problem and take inspiration from Machine Reading Comprehension (MRC) literature, to design our model. Our objective in designing such a model, is to exploit the local linguistic context in clinical text and enable intra-sequence interaction, in order to jointly learn to classify drug and disease entities, and to extract adverse reactions caused by a given drug. Our model makes use of a self-attention mechanism to facilitate intra-sequence interaction in a text sequence.  This enables us to visualize and understand how the network makes use of the local and wider context for classification.
\end{abstract}

\section{INTRODUCTION}
Adverse reactions caused by drugs is a potentially life-threatening problem. Extracting such adverse effects is a non-trivial problem. It has attracted the interest of researchers and Health-care Providers. ADE (Adverse Drug Event) is an umbrella term that includes adverse reaction to drugs, unintended side effects and effects from discontinuation or overdose of prescribed drugs. 94\% of ADE cases are under-reported by the official systems used in the pharmaceutical industry \cite{adestats}. It is reported that approximately 7000 deaths \cite{underreport} were caused by ADE annually, in a study conducted in 2000. ADE are hard to discover because it manifests on certain group of people in certain circumstances, and it may take a long time to expose. Recent research in post-market drug surveillance \cite{adestats, pharmaco, socialmedia, adednn} make use of informal text from social media platforms like twitter. Our research is entirely focused on biomedical text extracted from PubMed reports \cite{adecorpus}.

The objective of ADE task is to identify drug and disease mentions in a sequence and possible ADE relations between them. Most of the prior work model this task as a entity and relation extraction problem. There are two broad categories of models used to solve this task. The first category uses traditional pipeline method which consists of two steps - Named Entity Recognition (NER) followed by relation classification. Most early works \cite{pipeline1, pipeline2} in this area, use pipeline models for ADE task. Pipeline models heavily rely on manual feature engineering. 

The second category of models are largely based on End-to-End Deep Neural Networks (DNN). Recent advances in deep learning have spawned diverse groups of neural network architectures tailored for Computer Vision, Natural Language Processing (NLP), Speech Recognition, etc,. The promise of deep learning is automatic learning of significant features from large data. The NLP community has adopted Recurrent Neural Networks (RNN) for processing large, unstructured, variable length text sequences. Miwa et al., \cite{treelstm} used Long Short Term Memory (LSTM) \cite{lstm} based RNNs for relation classification. Li et al., \cite{ff} used a feed forward neural network to jointly extracting drug-disease entity mentions and their ADE relations.\\ 

Li et al.'s work \cite{core}, which appears to be the state of the art in ADE task, employs a neural joint model to extract entities and their relations. They use a bidirectional LSTM based RNN for learning entity representations from text sequences. The output of this network is fed as input to another bidirectional LSTM RNN, which learns ADE representations. The LSTM  parameters are shared by both the networks. The ADE network processes Shortest Dependency Paths (SDPs) between possible entities, based on the dependency graph of the text sequence. Our approach makes use of several techniques employed in \cite{core}.\\

In this work, we present a simple sequence model for the combined task of Entity Recognition (ER) and Advese Drug Event (ADE) extraction. The design of our model is inspired by modern end-to-end sequence processing Recurrent Neural Network architectures. It is considerably simple and easier to train and extensible to similar tasks in NLP. The model makes use of a self-attention mechanism to facilitate intra-sequence interaction i.e., interaction between constituencies of a sequence. We postulate that the attention mechanism, widely employed by the deep learning community, could be an apt replacement for SDP-based methods used in \cite{core}, \cite{treelstm} and \cite{sdp1}. Additionally, we make use of the heat maps generated by the attention mechanism, to visualize and understand how the network makes use of the local linguistic context and global semantic context, for entity recognition and ADE extraction.

\begin{figure}
\centering     
\subfigure[Sequence Length Histogram]{\label{fig:data_seqlen}\includegraphics[scale=0.3]{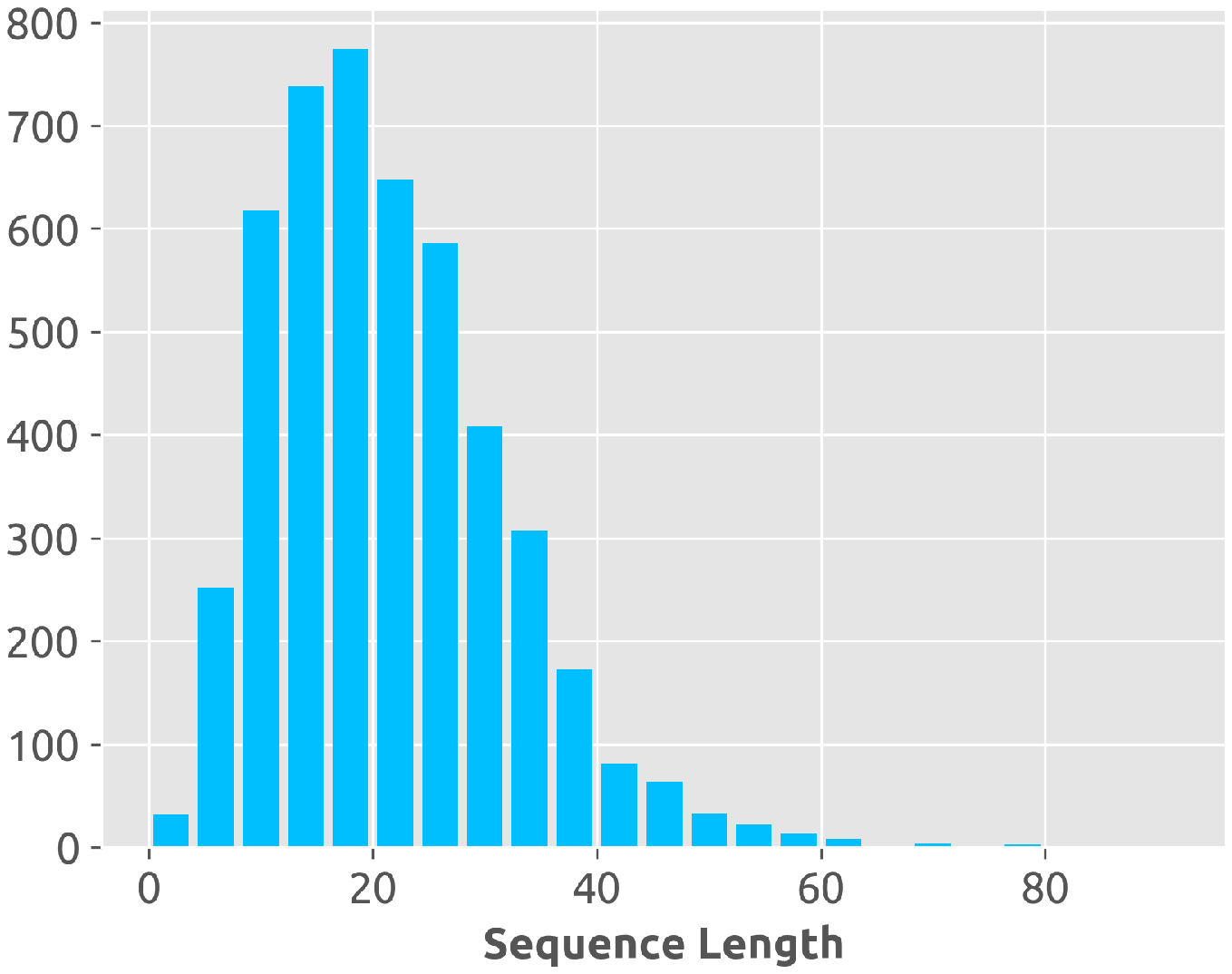}}%
\subfigure[Entity Count Histogram]{\label{fig:data_entities}\includegraphics[scale=0.3]{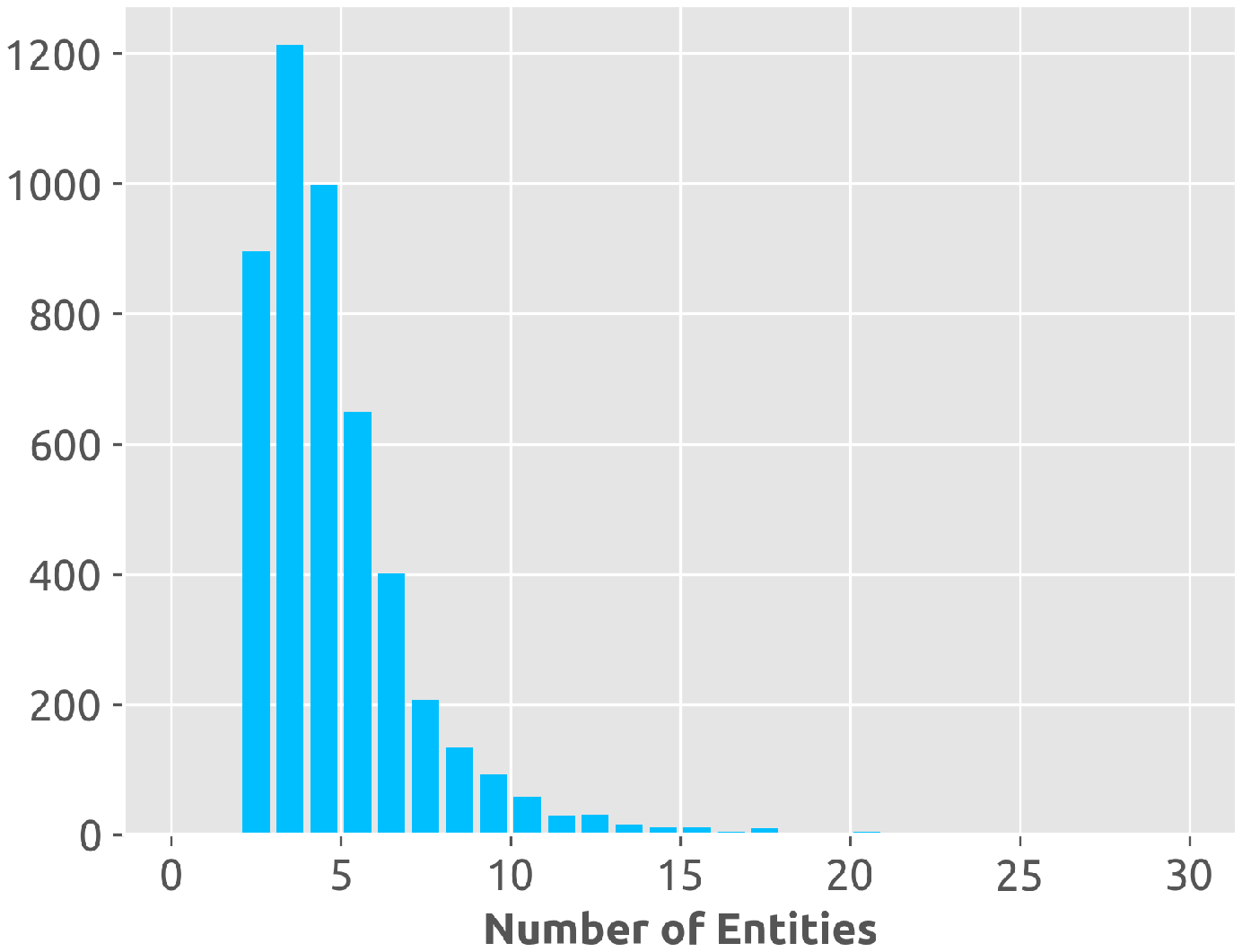}}
  \caption{ADE corpus statistics}
  \label{fig:data}
\end{figure}

\begin{figure*}
  \includegraphics[width=\textwidth]{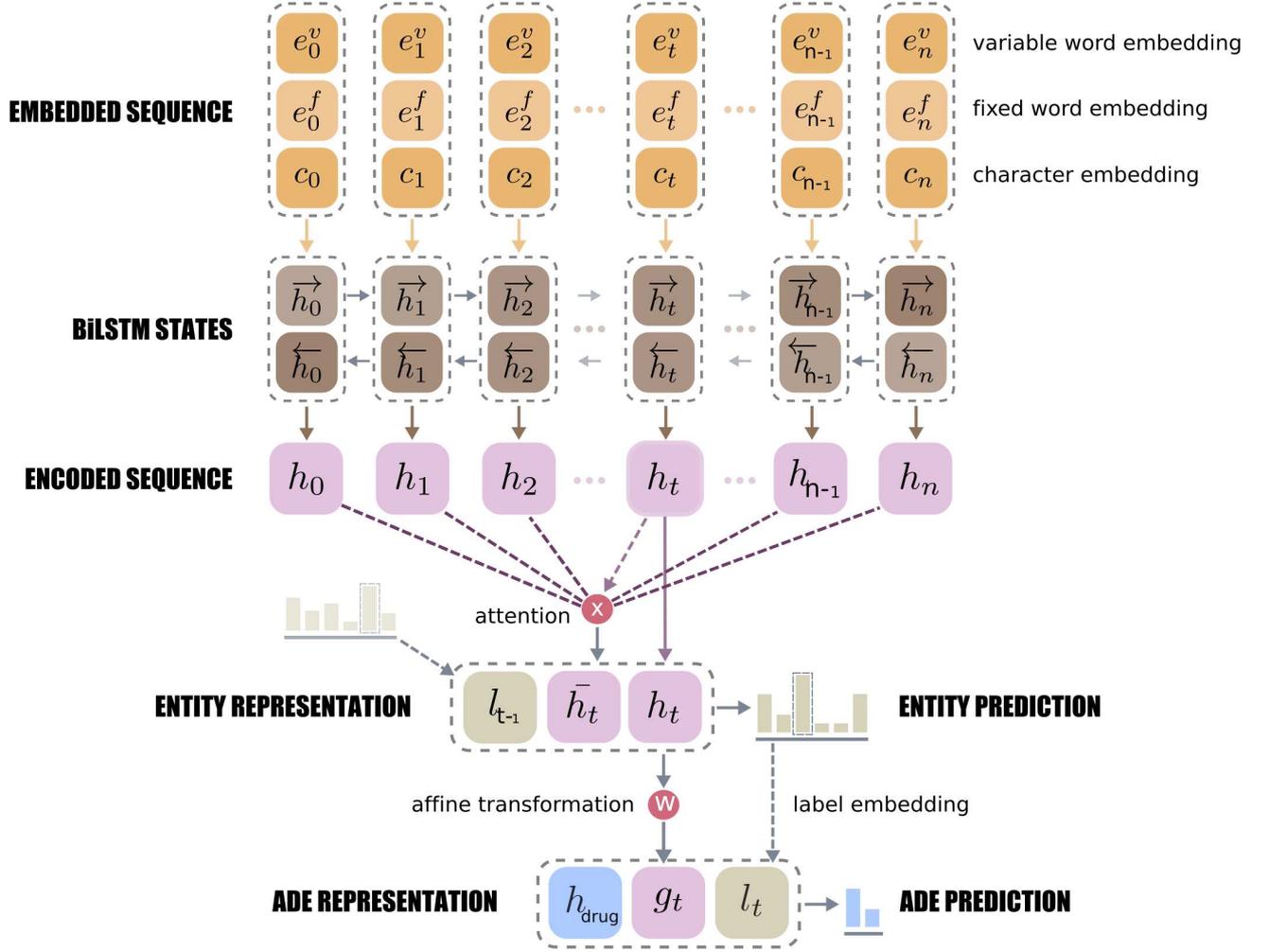}
  \caption{Architecture of Character-aware Attentive ADENet}
  \label{fig:arch}
\end{figure*}

\section{DATA}

The data consists of annotated sentences extracted from PubMed abstracts \cite{adecorpus}. 6821 sentences contain at least one ADE relation. 16695 unlabeled sentences with no ADE relations, are ignored. Each sample consists of an index and a text sequence, followed by a drug entity from the sequence and an ADE entity caused by the drug. We have modified the data samples such that each sample is annotated with one drug and a list of ADE's caused by that drug. This allows us to condition the model on a given drug and focus on the segments of text corresponding to the effect of the drug. As suggested in prior works \cite{overlapent, core}, 120 sentences with overlapping entities (e.g., "lithium intoxication", where "lithium" is a drug that causes "lithium intoxication") are removed from the dataset. We present the sequence length and average entity count in a sequence, as histograms in figure \ref{fig:data}.

\section{PROBLEM DEFINITION}

The problem we consider is two-fold. Given a sequence, task one is to recognize drug and disease entities in the sequence i.e., Entity Recognition. Task two is identifying adverse drug events in the text sequence i.e., ADE Extraction. Task one is a simple sequence labeling problem, where we predict the category of each token in the sequence. 

\begin{equation}p(y^{entity}_i | X_{seq}) = f_1(x_i, X_{seq})\end{equation}

We have considered task two as a question answering problem. In question answering, we are given (context, query) tuples and tasked to select an answer from a vocabulary. Most architectures in reading comprehension literature construct a query-aware context representation for selecting answers to query. Based on this idea, we considered ADE extraction as a question answering problem, where the text sequence becomes the context and the drug whose adverse effects are to be predicted, becomes the query. Rather than selecting an answer (adverse effect) from a vocabulary, we consider each token in the sequence as a potential ADE. We pose the question, "Is the $t^{th}$ word in the sequence an adverse effect of the given drug?". Thus we have loosened the constraints, to change a relation extraction problem, to a sequential binary classification problem.

\begin{equation}p(y^{ADE}_i | X_{seq}, X_{drug}) = f_2(x_i, X_{seq}, X_{drug})\end{equation}

\section{MODEL DEFINITION}

Our model can be partitioned into 4 stages - \textit{embedding}, \textit{encoding}, \textit{interaction} and \textit{prediction}. We use pretrained word embeddings to obtain distributed, continuous representation of tokenized sequences. Character-level word representations are obtained by using a combination of different sized convolutional filters, similar to Kim et al.'s work \cite{cnlm}. The word-level and character-level representations are concatenated and fed as input to the encoder. The encoder is a bi-directional LSTM, which processes the embedded sequence in either directions. The forward and backward hidden states of the LSTM are concatenated together to get an encoded sequence $\{[\overrightarrow{h_i}, \overleftarrow{h_i}]\}$. We use the same encoder to obtain an encoded representation of the drug. In the interaction stage, we iterate through the timesteps of the sequence and obtain a weighted representation of the whole encoded sequence, conditioned on current state. We combine the weighted representation with the current state to predict the entity type of the token corresponding to current state. We apply an affine transformation to the combined representation and then combine it with the encoded drug, to predict if the token corresponding to current state is an ADE of the given drug. Each stage is discussed in detail in the following sections.

\subsection{Embedding} \label{embsec}

Distributed representation of words introduced by Bengio et al., \cite{emb} has become competent a  replacement for the traditional bag-of-words encoding technique. Similar to \cite{core}, we adopt multiple embedding matrices to represent different representations of text sequence. Word-level and Character-level representations of input sequence are obtained by techniques explained in sections \ref{wemb} and \ref{chemb} respectively. In addition to these, we make use of Parts of Speech (PoS) and entity label information, by creating PoS embeddings and label embeddings. Both of these embedding matrices are randomly initialized.

\subsubsection{Word Embedding} \label{wemb}
We use two instances of pretrained word embeddings from PubMed word2vec \cite{word2vec}. We disable gradient updates on the first instance - \textit{Fixed Embedding} $(emb_f)$ while we enable gradient updates on the second instance - \textit{Variable Embedding} $(emb_v)$. Fixed embeddings provide good, stable representations for drug and disease entities in the sequence. Variable embeddings learn representations for common words and phrases in the corpus that highlight their role as linguistic supporting structures. We concatenate the  fixed and variable embeddings for word-level representation of sequence tokens.

\begin{equation}emb(x) = [emb_f(x); emb_v(x)]\end{equation}

\begin{table}
\begin{center}
\caption{Architecture of CharCNN}
  \begin{tabular}{ll}
    \toprule
	   {Filter Attributes} & {Values} \\
	         \midrule
       {Embedding Dimensions} & {25} \\
       {Filter widths ($w$)} & {[1,2,3,4,5,6]} \\       
       {Number of filter matrices} & {[25.$w$]} \\

    \bottomrule
  \end{tabular}
\label{tb:t0}
\end{center}
\end{table}

\subsubsection{Character-level Embedding} \label{chemb}
Convolutional Neural Networks (CNNs) have been shown to perform well on NLP tasks. \cite{cnlm} and \cite{tscratch} have used CNNs for learning character-level features in text. Kim et al., \cite{cnlm} propose a character-aware neural language model, that learns character-level word representations using CNNs - \textit{CharCNN}. They employ multiple convolutional filters of varying widths to obtain feature vectors for words in the sequence. We follow the same mechanism for character embedding.

Initially the text sequence is split into a nested sequence of characters. A lookup of character embeddings is performed and the vectors corresponding to individual characters in a word, are stacked together to form a matrix. This is followed by a series of convolutions with multiple filters of different widths. Max-over-time pooling is applied to the output of convolutional layer, to obtain character-level word embeddings for each word in the sequence. The details of filter size, number of filters and embedding size are tabulated in Table \ref{tb:t0}.

\subsection{Encoder}
The embedded text sequence which is comprised of word embedding (fixed and variable), character embedding and PoS feature embedding, is processed by the encoder, a bidirectional LSTM based RNN. The forward $\{\overrightarrow{h}_i\}$ and backward $\{\overleftarrow{h}_i\}$ hidden states of the RNN are concatenated together. Now we have a primary representation of the text sequence - the encoded sequence $\{h_i\}$. The drug, in it's embedded form, is also processed by the same encoder network. The final state of the RNN is considered the encoded drug representation $h_{drug}$.

\begin{equation}h_t = [\overrightarrow{h}_t; \overleftarrow{h}_t]\end{equation}

\begin{equation}h_{drug} = [\overrightarrow{h}_{drug}; \overleftarrow{h}_{drug}]\end{equation}

\subsection{Interaction Layer}\label{interaction}

During each step $t$ of encoder, information from the local neighborhood flows from either directions. The hidden state $\bar{h}_t$ represents the local context corresponding to $t^{th}$ time step. In \cite{core}, ADE relation extraction is done by building SDPs between drug entities and potential ADE candidates in the sequence. We have designed the interaction layer as an alternative to this approach. The objective is to facilitate interaction between different parts of the sequence. Our model considers each token in the sequence as a potential entity and an ADE candidate.

In the \textit{interaction} stage, we iterate through the encoded sequence and learn a entity representation and an ADE representation for each sequence. During each time step $t$, the model creates a weighted representation over all the encoded states $\{\bar{h}_i\}$, conditioned on the current state $h_t$. We choose multiplicative attention mechanism proposed in \cite{mulatt}, over additive attention \cite{addatt}, since the former is faster and more space-efficient \cite{att}.

\begin{equation}\bar{h}_i = \sum_j a_{ij}h_j\end{equation}
\begin{equation}\label{eq:att}a_i = softmax(f_{att}(h_i, h_j))\end{equation}
\begin{equation}f_{att}(h_i, h_j) = h_i^TW_ah_j\end{equation}

\subsection{Prediction}

At each step $t$ of the interaction layer, we construct entity and ADE representations corresponding to token $t$. For entity representation, we consider the embedding of predicted label of the previous token, $l_{t-1}$. We concatenate previous label embedding $l_{t-1}$ with weighted representation of encoded sequence $\bar{h}_t$ and the current state $h_t$. This gives us the entity representation $[l_{t-1}; \bar{h}_t; h_t]$. A $tanh$ non-linearity is applied, followed by an affine transformation. The output of this operation is normalized with softmax, which provides a probability distribution over the entity labels.

\begin{equation}p(y^{ent}_t | X_{seq}) = softmax(tanh(W^{ent}r^{ent} + b^{ent}))\end{equation}
\begin{equation}r^{ent} = [ l_{t-1}; \bar{h}_t; h_t]\end{equation}

We apply an affine transformation over $r^{ent}$ to get $\bar{r}^{ent}$ to enable variation between entity and ADE representation. We concatenate encoded drug state $h_{drug}$ with $\bar{r}^{ent}$ and the embedding of label predicted at current step $l_t$, to obtain the ADE representation, $r^{ade}$. By applying another affine transformation over $r^{ade}$, followed by softmax, we get the probability distribution over ADE labels, 0 or 1 corresponding to \textit{not an ADE} and \textit{ADE}. 

\begin{equation}p(y^{ade}_t | X_{seq}, X_{drug}) = softmax(W^{ade}r^{ade} + b^{ade})\end{equation}
\begin{equation}r^{ade} = [h_{drug}; \bar{r}^{ent}; l_t]\end{equation}
\begin{equation}\bar{r}^{ent} = W^cr^{ent} + b^c\end{equation}
\begin{equation}l_t = emb_l(\hat{y}_t)\end{equation}
\begin{equation}\hat{y}_t = argmax(p(y_t^{ent}| X_{seq}))\end{equation}

\begin{table}
\caption{Comparison with the state of the art}
  \begin{tabular}{lSSSSSS}
    \toprule
    \multirow{2}{*}{Model} &
      \multicolumn{3}{c}{Entity Recognition} &
      \multicolumn{3}{c}{ADE Extraction} \\
      & {P} & {R} & {F1} & {P} & {R} & {F1} \\
      \midrule
    Li \cite{core} & 82.70 & 86.70 & 84.60 & 67.50 & 75.80 & 71.40 \\
    Our Model & 88.41 & 82.41 & 85.30 & 86.28 & 87.29 & 86.78 \\
    \bottomrule
  \end{tabular}
\label{tb:t1}
\end{table}

\begin{table*}[ht]
\caption{Performance Comparison of model augmented with various features}
  \begin{tabular*}{\textwidth}{p{8cm}p{1cm}p{1cm}p{1cm}p{1cm}p{1cm}p{1cm}}
    \toprule
    \multirow{2}{*}{Model} &
      \multicolumn{3}{c}{Entity Recognition} &
      \multicolumn{3}{c}{ADE Extraction} \\
      & {P} & {R} & {F1} & {P} & {R} & {F1} \\
      \midrule
Baseline & 79.91 & 67.62 & 72.25 & 75.30 & 80.27 & 77.70\\
Baseline + PoS Features & 79.93 & 68.73 & 73.90 & 76.76 & 81.14 & 78.89\\
Baseline + Attention & 83.57 & 74.29 & 78.65 & 81.14 & 81.49 & 81.31\\
Baseline + Attention + PoS Features & 84.23 & 75.35 & 79.53 & 81.93 & 83.14 & 82.53\\
Baseline + Attention + Character Embedding & 82.69 & 77.74 & 80.13 & 79.35 & 86.20 & 82.62\\
Baseline + Attention + Character Embedding + PoS Features & 88.41 & 82.41 & 85.30 & 86.28 & 87.29 & 86.78 \\
    \bottomrule
  \end{tabular*}
  \label{tb:t2}
\end{table*}

\section{Model Variants and Features}
We have experimented with different variants of the model architecture presented in figure \ref{fig:arch}. The performance of these models are tabulated in Table \ref{tb:t2}. We started with a baseline model - \textbf{ADENet}, which consists of an LSTM-based bidirectional encoder. The final state of encoder which processes the drug, is taken as encoded drug representation. At each step of encoding the sequence, the hidden state $\bar{h}_t$ is combined with the embedding of entity label $\hat{y}^{entity}_{t-1}$ predicted in the previous step. This becomes the entity representation corresponding to token $X^{seq}_t$. An affine transformation, followed by a $tanh$ non-linearity gives the logits for entity classification. By combining the current hidden state $\bar{h}_t$ with label embedding of current entity prediction $\hat{y}^{entity}_t$ and the encoded drug $\bar{h}^{drug}$, we obtain the ADE representation of $X^{seq}_t$. Likelihood of ADE is obtained by applying softmax over affine transformed ADE representation.\\

Parts of speech tags of the text sequence are obtained using \textit{nltk}'s \cite{nltk} PoS tagger. PoS embeddings of the sequence are concatenated with the word embeddings to get a richer primary sequence representation. This model - \textit{Baseline + PoS Features} outperforms the baseline by a small margin.\\

The Interaction layer, defined in section \ref{interaction}, is added to the baseline - \textit{Baseline + Attention}. We can observe a sharp increase in the performance of this model, in both the tasks, compared to the baseline. This is due to the fact that during each step of prediction, the model has access to the encoded representation of the whole sequence. By conditioning the weighted representation on the current state, the model performs a search on the sequence, looking for information relevant to current prediction. The hidden state bottleneck \cite{addatt} is overcome by allowing the model to peek at the whole sequence. 

When PoS features are added to this model - \textit{Baseline + Attention + PoS Features}, there is an increase in performance on both the tasks. We theorize that the attention mechanism learns to make better use of new information provided by the additional features (PoS). This theory holds true when we add character embeddings to the model.

We then introduce character embedding as an additional feature to the model - \textit{Baseline + Attention + Character Embedding}. This results in a much better performance. The character embeddings contain rich morphological information, which are crucial in differentiating a drug or disease entity from a regular English word. 

And finally when we add PoS features to this model - \textit{Baseline + Attention + Character Embedding + PoS Features}, we obtain our best performing model by far (illustrated in figure \ref{fig:arch}). Based on these observations, we add that by including domain-specific or context-specific features and facilitating interaction between the features corresponding to individual tokens in the sequence, the performance on NLP tasks could be significantly improved.

\begin{figure}\label{fig:analysis}
\centering     
\subfigure[ER F1 Scores]{\label{fig:analysis_a}\includegraphics[scale=0.3]{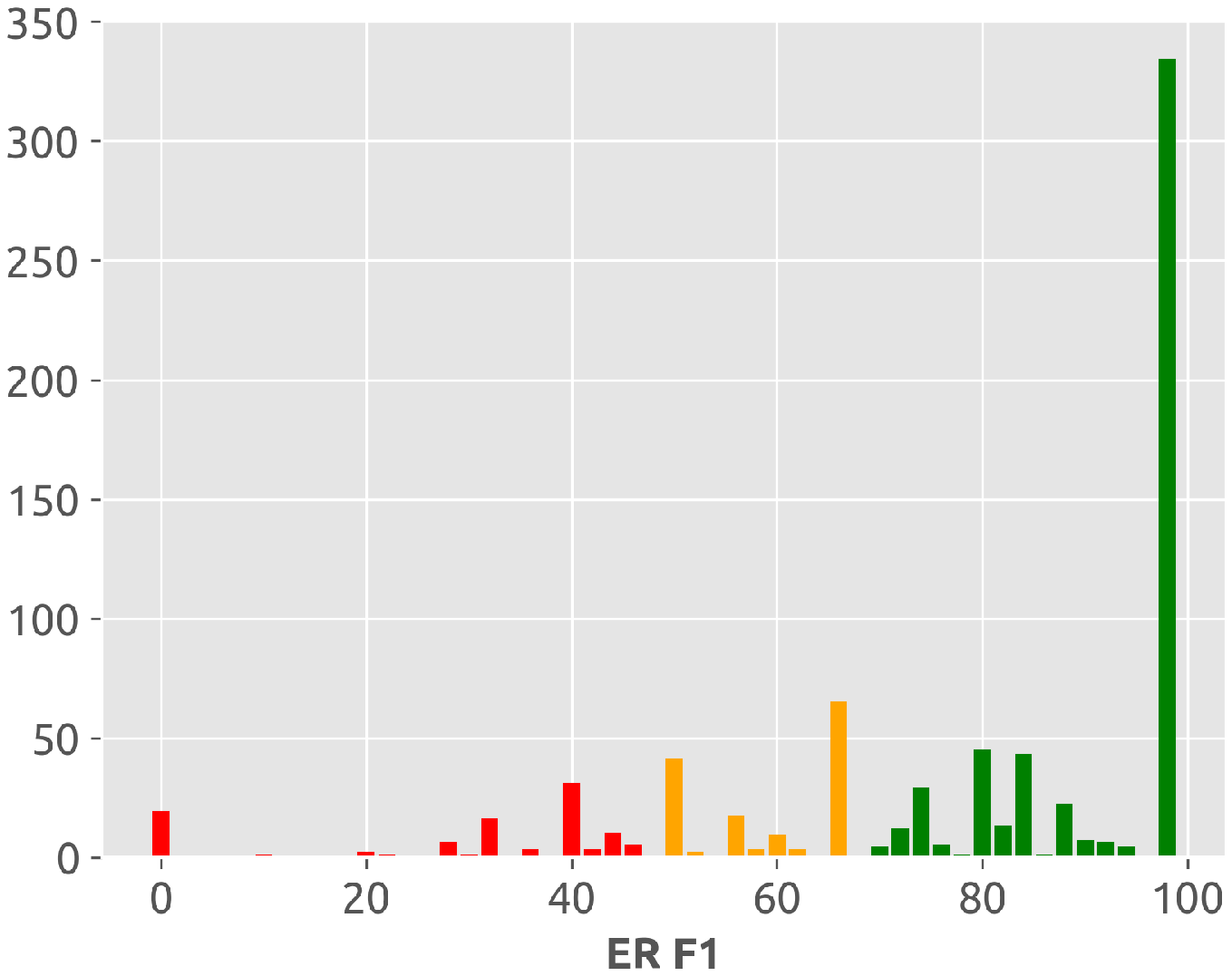}}%
\subfigure[ER F1 Scores $<$ 90.]{\label{fig:analysis_b}\includegraphics[scale=0.3]{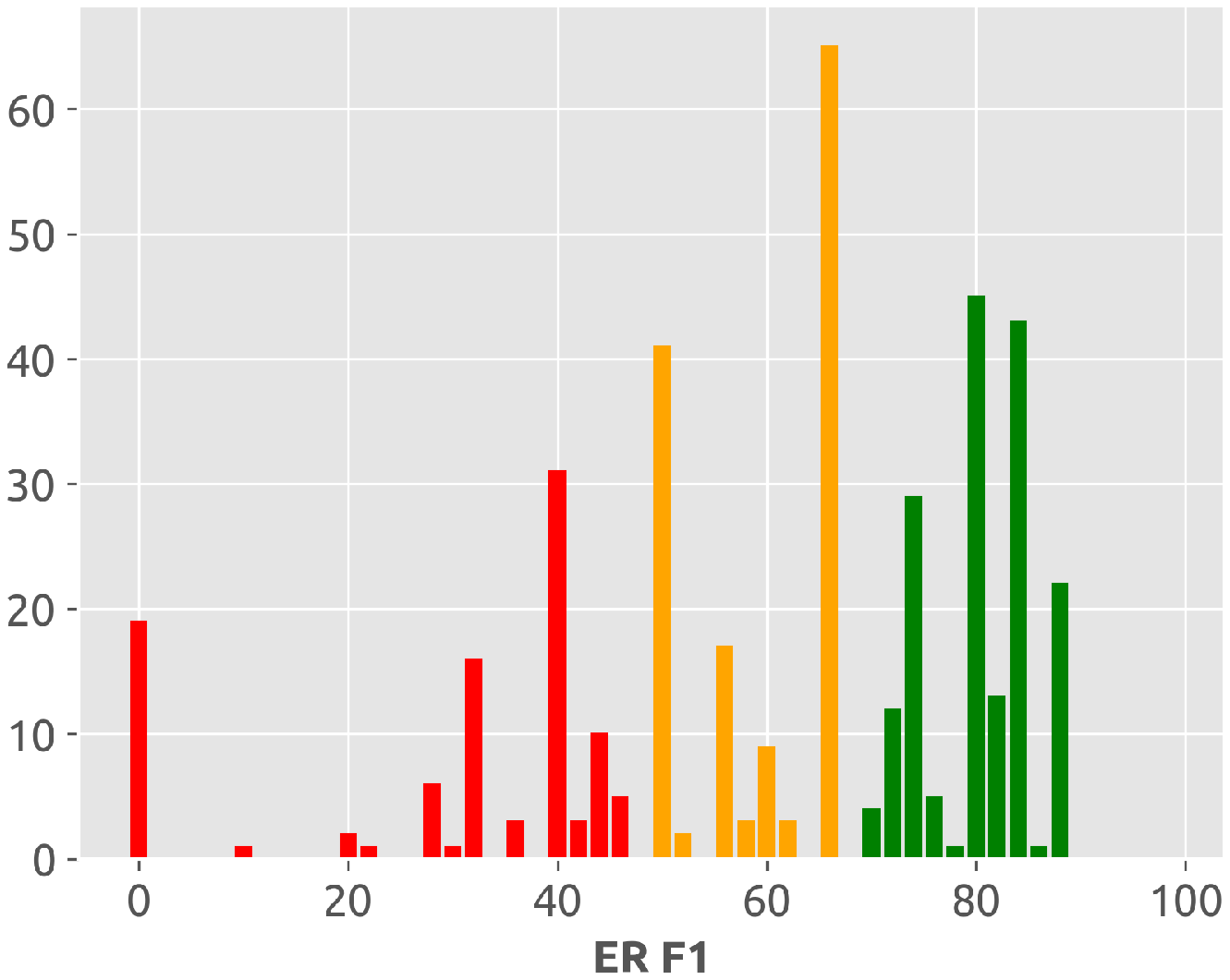}}
\subfigure[ADE F1 Scores]{\label{fig:analysis_c}\includegraphics[scale=0.3]{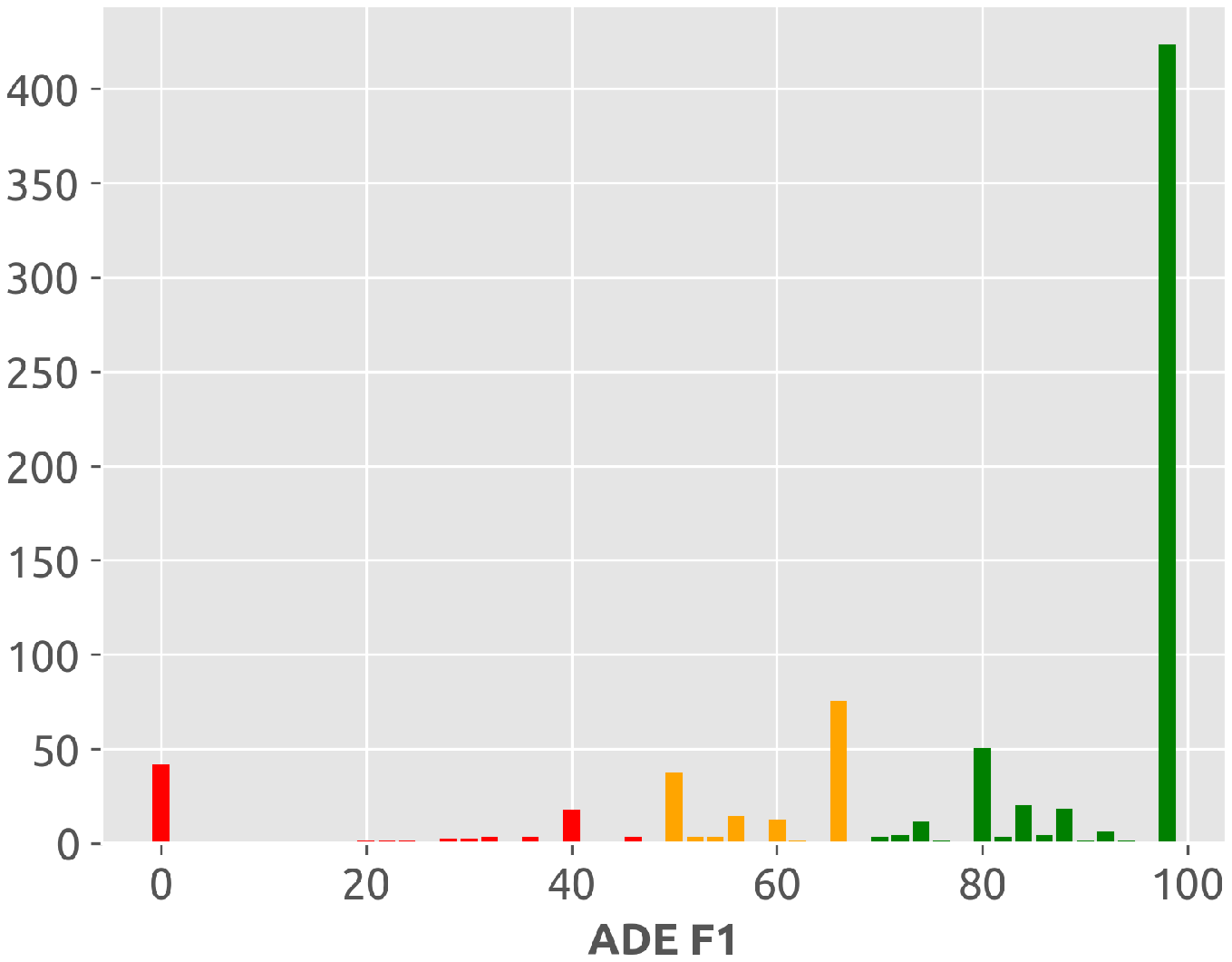}}%
\subfigure[ADE F1 Scores $<$ 90.]{\label{fig:analysis_d}\includegraphics[scale=0.3]{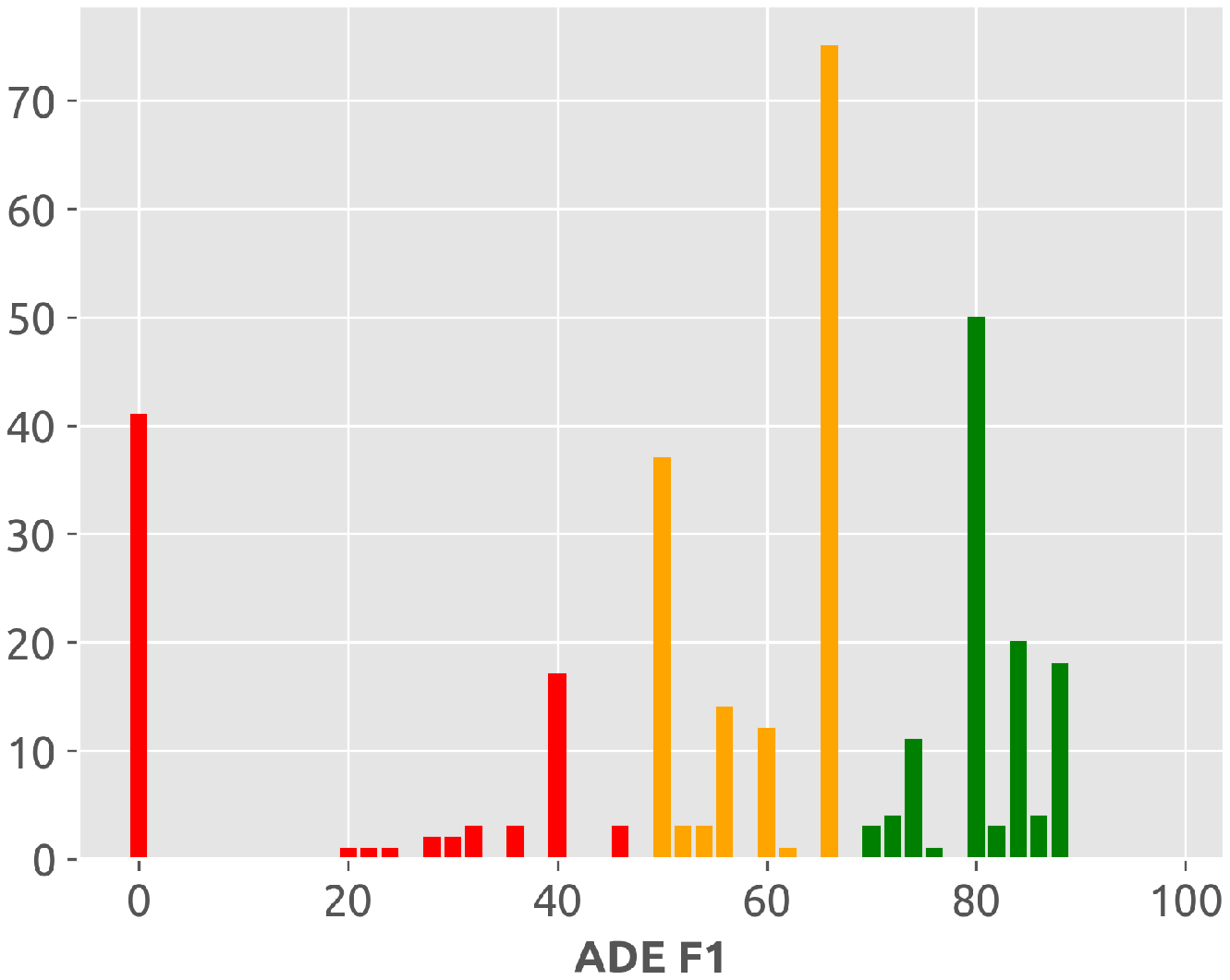}}
\caption{Histogram of F1 scores in Entity Recognition and ADE Extraction tasks are presented in figures (a) and (c) respectively. (b) and (d) ignore F1 scores greater than 90.}
\label{fig:analysis}
\end{figure}

\begin{figure}
  \centering
  \includegraphics[scale=0.5]{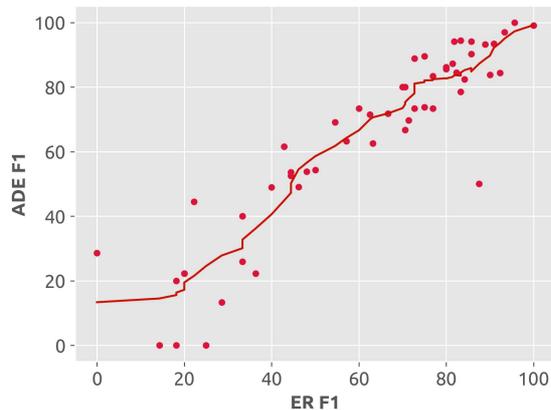}
  \caption{Correlation between F1 scores of ER task and ADE task}
  \label{fig:erade}
\end{figure}

\section{TRAINING}

We train our models with an initial learning rate of 0.001 and a batch size of 16, with Adagrad Optimizer \cite{adagrad}. We have split the dataset in 8:1:1 ratio into training set, test set and validation set. We use a dropout \cite{dropout} of $0.5$ for the LSTM encoder. All the trainable parameters of the model are initialized with a random uniform initializer within range (-0.01, 0.01).  All our models converge within 10 epochs. The dimensions of pretrained PubMed word embeddings is 200. Dimensions of character embedding, PoS embedding and label embedding, along with the rest of the hyperparameter settings, are listed in Table \ref{tb:t3}.

\section{EXPERIMENT RESULTS}
Table \ref{tb:t1} compares the performance of our best performing model with that of Li et al.'s state of the art model \cite{core}. Our model's performance in Entity Recognition is slightly higher compared to the state of the art. We also present the results of our ADE extraction system. It is unfair to compare it's performance with that of \cite{core}, as ADE extraction is treated as a relation extraction problem in the latter. Although, in a practical setting, we could augment an end-to-end neural network with a drug lexicon, to condition the model to extract ADE's caused by a specific drug. By constraining our model to focus on ADE's only relevant to a particular drug, we have achieved an F1 score of $86.78$, which is an improvement of approximately $15\%$ over the state of the art.

\begin{table}
\begin{center}
\caption{Hyperparameter Settings}
  \begin{tabular}{ll}
    \toprule
	   {Hyperparameter} & {Value} \\
	         \midrule

       {Word Embedding} & {400} \\
       {Character Embedding} & {525} \\       
       {PoS Embedding} & {25} \\
       {Label Embedding} & {25} \\       
       {BiLSTM hidden dimensions} & {[150, 150]} \\              
       {Learning rate} & {0.01} \\       

    \bottomrule
  \end{tabular}
\label{tb:t3}
\end{center}
\end{table}

\begin{figure*}
  \includegraphics[width=\textwidth]{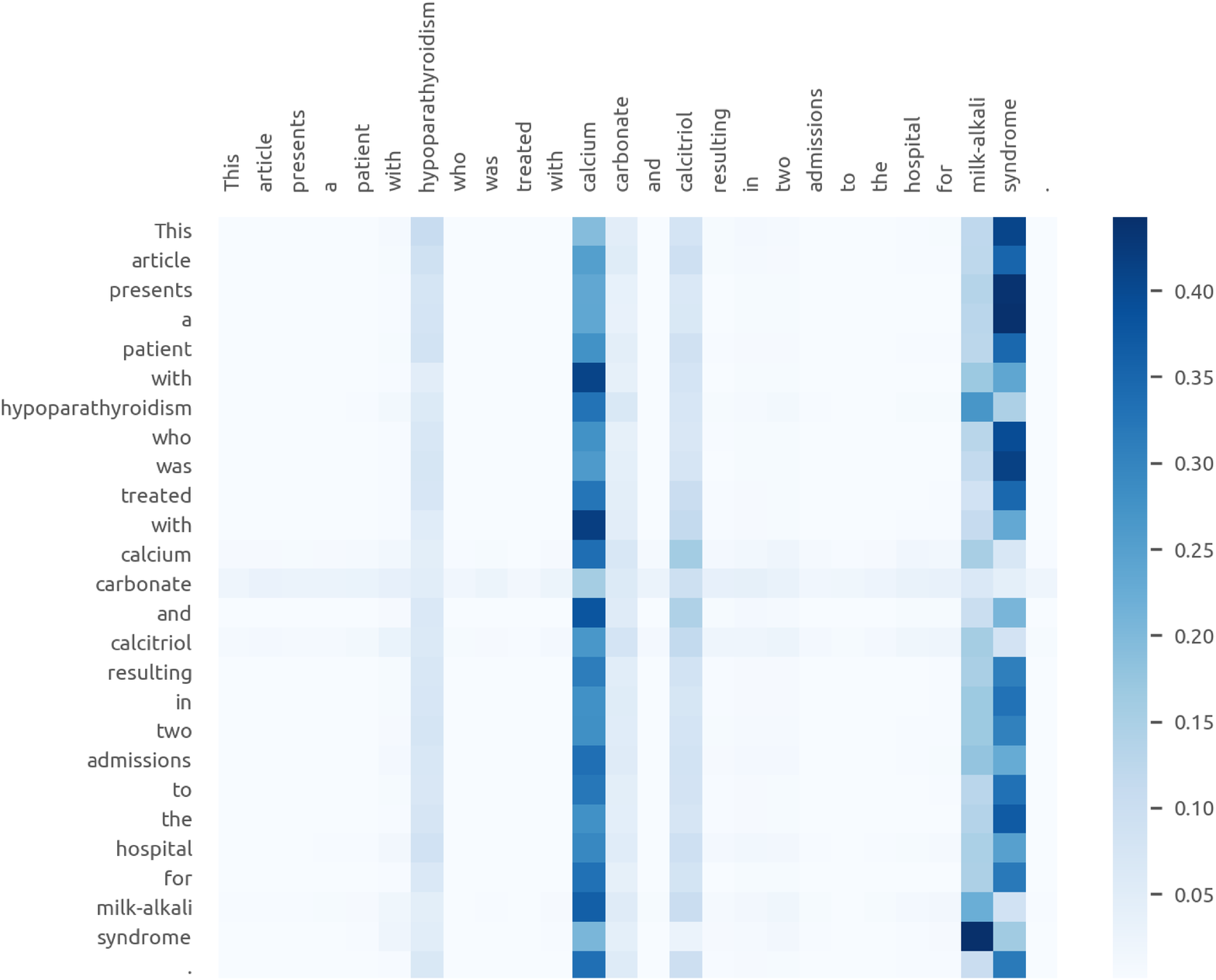}
  \caption{Visualization of Attention scores. To understand the figure, the sequence on the left should be read from top to bottom. Every row is a probability distribution (normalized attention score) across the whole sequence conditioned on the token on the left. Notice the peak in attention in positions corresponding to drug or disease entities.}
  \label{fig:heatmap}
\end{figure*}

\section{ANALYSIS}
Our analysis is focused on predictions made by our best performing model which includes Character Embedding, PoS features and Attention mechanism. We have clustered the data samples into three categories based on our model's performance on the sample. We studied the properties of each category of samples and present our observations in this section. The histograms of F1 scores of Entity Recognition and ADE extraction tasks are presented in figure \ref{fig:analysis}.  

We can clearly observe that more than 60\% of samples achieve an F1 score greater than 95 in ADE extraction task and almost 50\% of samples achieve an F1 score greater than 95 in Entity Recognition task. Figures \ref{fig:analysis_b} and \ref{fig:analysis_d} provide a closer look at the rest of the samples. The performance of the rest of the samples seem to be almost evenly distributed between scores of 20 and 90, with a few outliers. We can also notice a sharp peak around the score of zero.\\ 

We further investigate these samples by studying the effect of text sequence length, number of entities in the sequence and number of ADEs in the sequence, on performance. Based on which we conclude that there is no observable correlation between sequence length and performance. A similar comparison based on number of entities and number of drugs in the sequence, does not reveal any significant distinguishing features that separate good samples from bad samples. We then, compare the performance of our model in individual tasks. It is clear from figure \ref{fig:erade} that there is an increase in ADE performance with increase in ER performance. This kind of correlation is to be expected in a joint model, where sub-systems designed to solve different problems share parameters.\\

As discussed in section \ref{interaction}, we employ an attention mechanism across all the encoded states, corresponding to word tokens in the text sequences. This mechanism allows the system at each prediction step, to focus on information from the sequence relevant to current prediction. We have mapped the attention scores $a_i$ obtained from equation \ref{eq:att}, to the tokens in the text sequence and present a sample visualization of the interaction layer in figure \ref{fig:heatmap}. We can observe clear peaks in attention scores, corresponding to drug and disease entities. This pattern is observed in every good sample (high performance in both the tasks) from the test set. To add to that, in most bad samples, the attention seems to be scattered across the tokens in the sequence. This could be interpreted as the model searching the encoder states, for information that could be useful for prediction and failing to find it.

\section{CONCLUSION}

In this paper, we propose a joint model for Entity Recognition and Adverse Drug Event extraction in biomedical text. By loosening the constraints on ADE extraction, our model outperforms the state of the art by a large margin. The performance of our Entity Recognition sub-system is slightly better than that of state of the art. We observe that by using an attention mechanism to facilitate intra-sequence interaction, we could replace SDP based methods used in prior work. Additionally, the use of attention mechanism allows us to investigate the dependence of classifier on different segments of the text sequence.

\addtolength{\textheight}{-12cm}

\end{document}